\documentclass[letterpaper, 10 pt, conference]{ieeeconf}  

\IEEEoverridecommandlockouts     

\usepackage{graphicx,algorithm,algorithmic}
\usepackage{amsmath,amssymb,amstext,amsfonts,mathrsfs,mathtools,color}
\usepackage[algo2e,linesnumbered]{algorithm2e}
\usepackage{tikz}
\usepackage{setspace}
\usepackage{array} 
\usepackage{booktabs}  
\usepackage{bbm}
\usepackage{wasysym}
\usepackage{textcomp}
\usepackage[sort,compress]{cite}

\usepackage{longtable,tabularx}

\usepackage{subfig}
\usepackage{comment}
\usepackage{multirow}
\usepackage{multicol}
\usepackage{caption}
\usepackage{tabularx}
\usepackage{dblfloatfix}   
\usepackage{comment}
\usepackage{hyperref}

\allowdisplaybreaks 

\newcommand{\bluetext}[1]{{\leavevmode\color{blue}#1}}

\title{\LARGE \bf Robust Reinforcement Learning Algorithm for \\ Vision-based  Ship Landing of UAVs}

\author{Vishnu Saj, 
        Bochan Lee, 
        Dileep Kalathil, 
        and~Moble~Benedict
        \thanks{Corresponding author Vishnu Saj is with the Department
of Aerospace Engineering, Texas A\&M University, College Station,
TX, 77840 USA e-mail: {\tt vishnu02saj@tamu.edu}. This work has been funded by the Center for Unmanned Aircraft Systems (C-UAS), a National Science Foundation Industry/University Cooperative Research Center (I/UCRC) under NSF award Numbers IIP-1161036 and CNS-1946890, along with significant contributions from C-UAS industry members.}
}

\begin{document}

\maketitle
\thispagestyle{empty}
\pagestyle{empty}

\begin{abstract}
This paper addresses the problem of developing an  algorithm for autonomous ship landing of vertical take-off and landing (VTOL) capable unmanned aerial vehicles (UAVs), using only a monocular camera in the UAV for tracking and localization. Ship landing is a  challenging task   due to the small landing space, six degrees of freedom ship deck motion,  limited visual references for localization, and adversarial environmental conditions such as wind gusts. We first develop a computer vision algorithm which estimates the relative position of the UAV with respect to a horizon reference bar on the landing platform using the image stream from a monocular vision camera on the UAV. Our approach is motivated by the actual ship landing procedure followed by the Navy helicopter pilots in tracking the horizon reference bar as a visual cue. We then develop a robust reinforcement learning (RL) algorithm  for controlling the UAV towards the landing platform even in the presence of adversarial environmental conditions such as wind gusts. We demonstrate the superior performance of our algorithm compared to a benchmark nonlinear PID control approach, both in the simulation experiments using the Gazebo environment and in the real-world setting using  a Parrot ANAFI quad-rotor and sub-scale ship platform undergoing 6 degrees of freedom (DOF) deck motion. \textcolor{blue}{The video of the real-world experiments and demonstrations is available at this} \href{https://www.youtube.com/watch?v=4SiSVvzDrjg}{URL}.
\end{abstract}


\section{Introduction}

In recent years, there has been an increasing interest in developing autonomous control algorithms for UAVs for tracking a moving target and landing on it using only the camera sensor information for localization\cite{sanchez2014approach,truong2016vision,holmes2016autonomous,meng2019visual}. Developing such an autonomous algorithm for a vertical take-off and landing (VTOL) capable unmanned aerial vehicle (UAV) on a small moving ship, using only a monocular camera in the UAV for tracking and localization, is a particularly challenging task   due to the small landing space, six degrees of freedom ship deck motion,  limited visual references for localization, and adversarial wind gusts. The classical control approaches for this problem have been to use  proportional-derivative (PD) control \cite{daly2015coordinated, holmes2016autonomous}, proportional-integral-derivative (PID) control \cite{araar2017vision,truong2016vision},  linear quadratic regulator (LQR) \cite{lee2018helicopter,lee2020development,ghamry2016real}, adaptive control \cite{hu2015fast,kim2016landing,xia2020adaptive}, and model predictive control (MPC) \cite{vlantis2015quadrotor}. While these approaches are partially successful in addressing the problem,  they are often restricted to very specific settings due to three crucial weaknesses. First, the performance of the PD/PID type controllers solely depends on the design of its gain parameters,  which are typically difficult to fine-tune, especially in the simulator setting. PD/PID controllers have also limited transient response capabilities.  Second, the MPC and LQR approaches often require a very sophisticated analytical model of the real-world UAV and obtaining such a model can be very challenging in practice. Moreover, for computational tractability, the design of the optimal control policies using these approaches is often limited to simplified settings such as linear policy and quadratic costs. Third, the classical control approaches are typically  not robust against adversarial disturbances such as wind gusts and parameter mismatches/uncertainties between the simulator model and the real-world system model. These weaknesses often result in  unsatisfactory performance in real-world settings.   In this work, we propose a new autonomous control algorithm using  the robust reinforcement learning approach in order to overcome these challenges and  to achieve superior performance  both in simulations  and in real-world experiments.


 Reinforcement Learning (RL) is an area of machine learning that addresses the problem of learning the optimal control policy for a stochastic dynamical system when its model is unknown.  RL algorithms have seen impressive successes recently in a number of application such as playing games \cite{mnih2015human, silver2016mastering} and robotics   \cite{lillicrap2016continuous, levine2016end, akkaya2019solving, haarnoja2019learning}.  However, most of these successes are either in the simulation domain or in the structured real-world settings, which are  significantly different from the challenging real-world setting such as the VTOL problem in the presence of adversarial wind gusts. Training RL algorithms in the real-world setting is infeasible because it can be catastrophic; for example, an undertrained policy may crash the drone.  This challenge is typically overcome by training the RL controller in a simulator. However, it is very difficult to incorporate the diverse and complex real-world  drone-environment interactions  in a simulator. Moreover, the parameter mismatches between the simulator model and real-world environment  cannot be accounted while training the RL control policies in a simulator. This leads to the problem known as \textit{simulation-to-reality (sim-to-real) gap}, where the RL policies learned using a  simulator may not perform well for the real-world setting. In this work, we  overcome this challenge  by developing a problem specific robust RL control policy and architecture by adapting the domain randomization approach  \cite{sadeghi2017cad, tobin2017domain, peng2018sim} for the ship landing problem. 

One integral component of any  UAV algorithm capable of autonomous landing is a computer vision based algorithm for tracking the moving target and estimating the position of the UAV with respect to this target. Previous works used methods such as tracking the H/T-shaped landing marks,   points, or lights  on the deck \cite{sanchez2014approach, xu2009research,  meng2019visual,yakimenko2002unmanned}.  Methods involving visual tracking of deck motion is not ideal for VTOL  capable UAVs because actively controlling the UAV to match the complex deck motion could excite unstable UAV attitude dynamics. This is even more unsafe when the aircraft is in close proximity to the moving deck because even a small control error can cause a catastrophic accident due to an impact by the deck.  In this paper, we develop a computer vision based algorithm  inspired by the practical ship landing procedure that navy helicopter pilots follow \cite{nato, lumsden1998challenges, colwell2002maritime, stabhorBar}. This procedure uses a gyro-stabilized (indicating the true horizon independent of ship motions) \textit{horizon reference bar} as a visual cue.  The pilot stabilizes the helicopter attitude using this visual cue and then commands vertical landing, agnostic of the deck motion. We develop a computer vision algorithm which detects this horizon reference bar, and continuously estimates the relative position of the UAV with respect to this horizon reference bar using only a monocular vision camera on the UAV. 

Our main contribution are the following:
\begin{itemize}
    \item We develop a robust reinforcement learning based control algorithm for autonomous vision-based ship landing of VTOL capable UAVs.
    \item We develop a computer vision based algorithm for estimating the   relative position of the UAV with respect to the horizon reference bar using only a monocular vision camera. 
    \item We demonstrate the superior performance of our algorithm in a real-world setting using a Parrot ANAFI quad-rotor and sub-scale ship platform undergoing 6 degrees of freedom (DOF) deck motions. 
\end{itemize}



\textbf{Related work:}  RL-based flight attitude control is discussed in \cite{RL1}. An RL approach for  UAV landing task on a moving platform using a variant of the DDPG algorithm is discussed in \cite{RLland}. An actor-critic RL framework used to fly a UAV by following designated waypoints is presented in \cite{RLtrack}. In \cite{RL2}, a variant of the DDPG algorithm is used to recover a UAV attitude quickly from an out-of-trim flight state. By leveraging a probabilistic model of drone dynamics, a model-based reinforcement learning strategy is used to control a quadrotor in \cite{RL3}. Using an imitate-reinforce training framework, \cite{lin2019flying} proposed an end-to-end policy network for enabling a drone to fly through a tilted narrow gap.\cite{lambert2019low} used  model-based reinforcement learning on a Crazyflie centimeter-scale quadrotor with rapid dynamics to predict and control on a low level. However, these works do not address the problem of designing RL controllers that are robust against adversarial disturbances such as wind gusts.

\section{Relative Position Estimation using Monocular Vision Camera on the UAV}

\begin{figure}[t]
\centering
\vspace{2mm}
\includegraphics[width=0.35\textwidth]{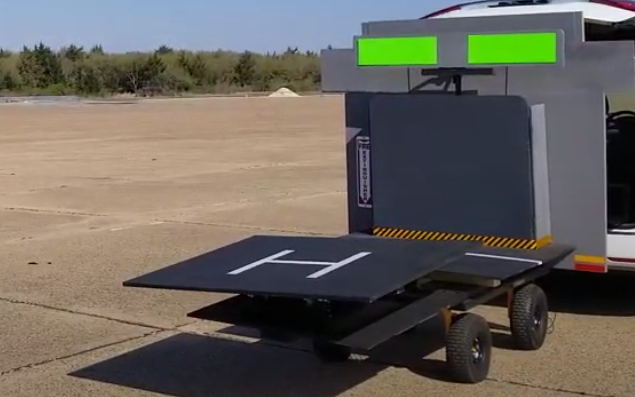}
\caption{Ship platform model with horizon reference bar and motion deck used in our real-world demonstrations.}
\label{fig:demo-model-ship-bar-deck}
\vspace{-0.4cm}
\end{figure}
\begin{figure}[h!]
\centering
 \includegraphics[width=0.44\textwidth]{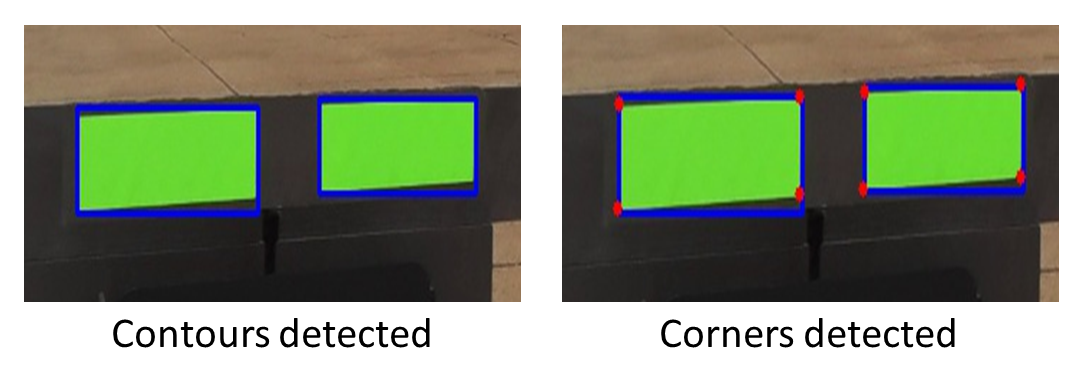}
\caption{Detecting the horizon reference bar.}
\label{fig:corner-detection}
\vspace{-0.6cm}
\end{figure}

We develop a computer vision algorithm which first detects the horizon reference bar on the ship platform and then estimates relative position  of the UAV with respect to this reference. The ship platform model with the horizon reference bar and the motion deck used in our real-world demonstration is shown in Fig.~\ref{fig:demo-model-ship-bar-deck}. As mentioned before, this approach of using  horizon reference bar as the primary visual cue is motivated  by the practical ship landing procedure that navy helicopter pilots follow \cite{stabhorBar,nato}. The relative position estimated by our computer vision algorithm is then used as part of the state information in the RL control algorithm.

A more detailed discussion of our  relative position estimation apporach is available in the technical report \cite{lee2022intelligent}, which includes the details omitted from here due to page limit.

\subsection{Image Filtering, Corner Detection and Screening}


Our algorithm takes the monocular camera image captured by the UAV as input, and first performs a Hue-Saturation-Value (HSV) filtering to detect  the green rectangles corresponding to the horizon reference bar.  In the HSV filtered image, there may  exist small white patches outside the rectangles and black voids inside the rectangles. We use the morphological opening technique  to remove the white patches in the image and the morphological closing is used to fill up small voids in those rectangles. We then use the watershed algorithm \cite{Couprie2003}  to obtain clear boundaries of the rectangles. 

Once the green rectangles are isolated, the next goal is to detect the contours and corner points. To detect the eight corners precisely, our algorithm first find the contours of the detected region and bound it in rectangles as shown in Fig.~\ref{fig:corner-detection}. Thus, the size and shape of the detected areas are very close to the green rectangles, and the corners of those bounding rectangles can be used as rough estimates of the actual corners. We adapt the F\"orstner corner detection \cite{forstner1987fast} method to detect the corner points of the rectangles  based on the rough corners obtained by contour detection.

The algorithm then uses a screening procedure  to ensure that no false corners are detected in the image. All the detected corners are sorted in a particular order,  and a simple coordinate system is used to identify the height,  width and the slope of different sides of the rectangles. Even though the detected regions are not perfect rectangles in the image, the width and height of the rectangles have similar lengths and slopes. A $\pm 10\%$ tolerance level is set for the lengths and a $\pm 5\%$ tolerance level is set for the slopes.

\subsection{Relative Position  Estimation}

\begin{figure}[h!]
\centering
\includegraphics[width=0.45\textwidth]{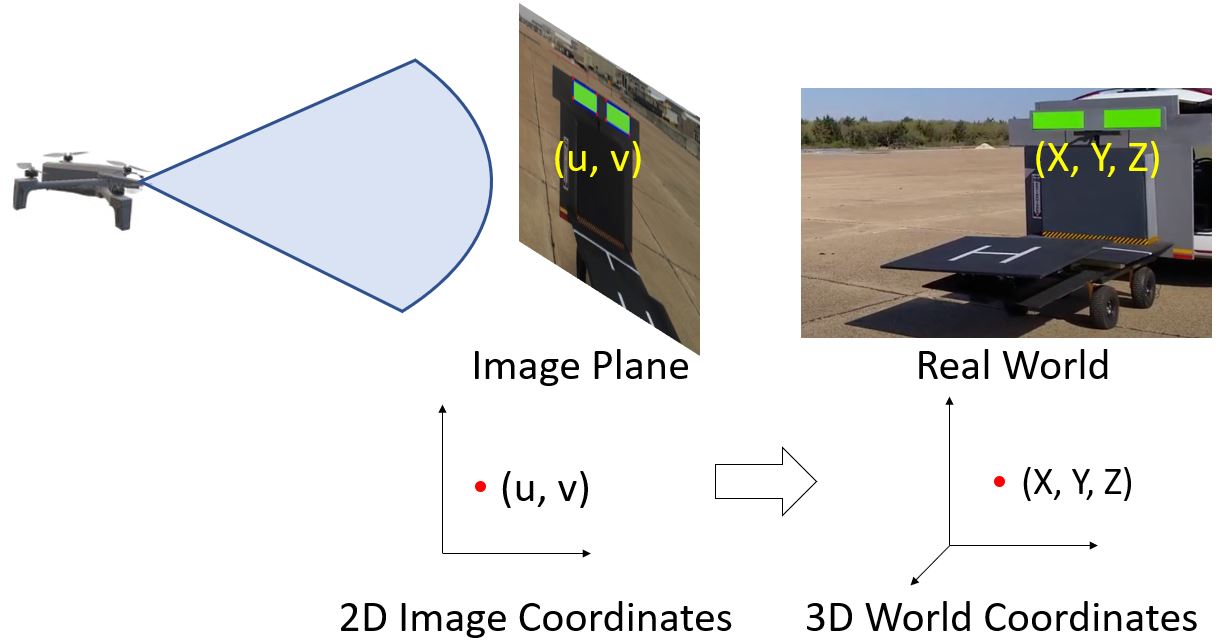}
\caption{Geometric relationship of 2D image coordinates and 3D real-world coordinates.}
\label{fig:geometric_relation}
\vspace{-0.5cm}
\end{figure}

Our estimation is based on a single camera calibration method using a planar object \cite{est1,est2} and  a conventional pinhole camera model is used to derive the geometric relation.  Let $(X, Y, Z)$ is a point on the visual cue and let $(u, v)$ is the corresponding point on the image frame (pixel position in the image), see Fig.~\ref{fig:geometric_relation}. Their correspondence is given by 
\begin{equation}
\hspace{-1.5cm}
\label{img_coord}
\hspace{2.0cm}
s
 \begin{bmatrix} u \\ v \\ 1 \end{bmatrix}
 =
 \boldsymbol{A}
  \begin{bmatrix}
   \boldsymbol{R} & t \\
   0 & 1\\
   \end{bmatrix}^{-1}
 \begin{bmatrix} X \\ Y \\ Z \\1 \end{bmatrix},
\end{equation}
where $s$ is the scaling factor, $A$ corresponds to the intrinsic camera parameters, $R$ is the rotation matrix and $t$ is the translation vector.  Given a set of $n$ 3D coordinates of an object and its corresponding 2D projections on the image,  the Perspective-n-Point (PnP) algorithm  can be used to determine the relative position and orientation. First, {the PnP} algorithm solves  Eq.\eqref{img_coord} to obtain the rotation matrix $R$ and the translation vector $t$. The PnP algorithm uses an iterative approach called Levenberg-Marquardt optimization \cite{lev1,lev2} to minimize the re-projection error. Once the  $R$ matrix and $t$ vector is computed,  the UAV camera position with respect to the visual cue can be obtained as $[X_{\text{cam}} ~ Y_{\text{cam}} ~ Z_{\text{cam}} ] = -\boldsymbol{R}^{-1}t$.

\section{Robust Reinforcement Learning based Control Algorithm}

\subsection{Reinforcement Learning Preliminaries}

We first give a brief overview of the Markov Decision Processes (MDP) and  reinforcement learning terminologies.



An MDP can be defined as a four-tuple $(\mathcal{S},\mathcal{A},P,R)$, where $\mathcal{S}$ is the state space and $\mathcal{A}$ is the action space, $P(s'|s,a)$ is the probability of transitioning from state $s$ to $s'$ upon taking action $a$, and $R(s,a)$ is the reward.  A control policy $\pi: \mathcal{S} \rightarrow \mathcal{A}$ specifies the control action to take in each possible state. The performance of a policy is measured using the metric of value of a policy, ${V}_{\pi}$, defined as ${V}_{\pi}(s) = \mathbb{E}_{\pi} [\sum^{\infty}_{t=0} \gamma^{t} R(s_{t}, a_{t}) | s_{0} = s ]$ where, $s_{t+1} \sim {P}(\cdot|s_t,a_t), a_t = \pi(s_t)$,  $s_t$ is the state of the system at time $t$, and $a_t$  is the action taken at time $t$, and $\gamma \in (0, 1)$ is the discount factor.  The goal is to find the optimal policy $\pi^{*}$ that achieves the maximum value, i.e, $\pi^{*} = \arg \max_{\pi} {V}_{\pi}$. The corresponding value function, ${V}^{*} = {V}_{\pi^{*}}$, is called the optimal value function. The optimal value function and policy  satisfy the Bellman equation, $\pi^*(s) = \arg \max_{a} (R(s) + \sum_{s' \in \mathcal{S}} {P}(s'|s,a) {V}^{*}(s'))$.

When the system model ${P}$ is known, the optimal policy  can be computed using dynamic programming. However, in most real-world applications, the  system model is either unknown or  difficult to estimate precisely. Even if the model is  known, directly computing the optimal nonlinear control policy is typically intractable for systems with large state and action spaces. RL offers a data driven and computationally tractable approach for learning the optimal control  policy using only the trajectory samples generated from an offline simulator of the system.

Policy gradient algorithms are  a popular class of RL algorithms for systems with continuous state and action spaces.  In a policy gradient algorithm, we represent the policy as $\pi_{\theta}$, where $\theta$ denotes parameters of the neural network that  represents the policy. Let $J(\theta) = \mathbb{E}_{s}[{V}_{\pi_{\theta}}(s)],$ where the expectation is w.r.t. to a given initial state distribution.  The goal is to find the optimal parameter $\theta^{*} = \arg \max_{\theta} J(\theta)$. This is achieved by implementing a gradient ascent update, $\theta_{k+1} = \theta_{k} + \alpha_{k} \nabla J(\theta_{k})$, where $\alpha_{k}$ is the learning rate. The gradient $\nabla J(\theta)$ is given by the celebrated policy gradient theorem as $\nabla J(\theta) = \mathbb{E}_{\pi_{\theta}}[{Q}_{\pi_{\theta}}(s,a) \nabla \log \pi_{\theta}(s,a)]$, where expectation is w.r.t. the state and action distribution realized by following the policy $\pi_{\theta}$. Here, ${Q}_{\pi_{\theta}}$ is the Q-value function corresponding to the policy $\pi_{\theta}$. The  Q-value function is also represented using a neural network (different from the one used for policy representation). There are many popular policy gradient algorithms such as TRPO \cite{schulman2015trust}, PPO \cite{schulman2017proximal}, SAC \cite{haarnoja2018soft}. In this work, we adapt the state-of-the-art twin delayed DDPG (TD3)  algorithm \cite{TD3} to develop a robust control policy for the VTOL problem.





\subsection{VTOL UAV Control as an RL Problem}

\subsubsection{Physical UAV and the Simulator Model} For the real-world demonstration, we use  Parrot Anafi \cite{parrot}, a commercial off-the-shelf quadrotor UAV. This particular UAV comes with a Gazebo simulator model  which can be used for training the RL algorithm.  Gazebo is a realistic simulation engine that is widely used for robotics applications  \cite{gazebo}. The RL algorithm  can communicate with the simulation engine using a framework called Olympe \cite{olympe}. Olympe allows the control of UAV and access to its sensors through python scripts.

\subsubsection{State Space} The relative position of the UAV w.r.t to the target (horizon reference bar) is used as part of the state.   We also include the velocity of the UAV as a part of the state. The velocity can be obtained either directly from the corresponding sensor or simply by a numerical calculation from the current and past position states. We use the past five position and velocity measurements as the current state. More precisely, $s_{t} = (p_{t}, v_{t}, p_{t-1}, v_{t-1}, \ldots, p_{t-5}, v_{t-5})$, where $p_{t}$ and $v_{t}$ are the relative position and velocity of the UAV w.r.t. the target in Euclidean coordinates.




\subsubsection{Action Space} The Parrot Anafi drone we use  has four different control actions: roll, pitch, yaw, and heave. During the experiments, we observed that the optimal roll and pitch actions heavily depend on the wind disturbances. Also, the main objective here is vertical landing. So, we consider only the  roll  (to move the UAV  right or left) and the pitch  (to move the UAV  forward or backward) as the  actions for the RL controller. For the Parrot Anafi drone, the roll and pitch action can  be controlled independently. The roll controller objective is to achieve a certain target on the roll axis and maintain that position without considering the pitch motion. Similarly, for the pitch controller, the objective is to maintain a position on the pitch axis independent of the roll motion.

\subsubsection{Reward Function} Designing the appropriate reward function that implicitly represents the underlying real-world task is one of the most important aspect for developing an RL algorithm for that real-world task. We use a carefully designed a reward function for our problem as given below.
\vspace{-4mm}
\begin{align*}  
\label{eq:reward}
\text{Reward} = \left\{ \begin{tabular}{ll}
    \vspace{-1mm}
    \multirow{2}{*}{$-\frac{1}{20}|a_{\text{diff}}|-\frac{1}{10}|a|$}  & \text{if}~  $|d| \leq 0.1$\\
     \vspace{0.2cm}
     & (Region-1)\\
    \vspace{-1mm}
    \multirow{2}{*}{$-2|d|-\frac{1}{20}|a_{\text{diff}}|-\frac{1}{10}|a|$}  & \text{if}~  $0.1 \leq |d| \leq 0.4$\\
    \vspace{0.2cm}
    & (Region-2)\\
    \vspace{-1mm}
     \multirow{2}{*}{$-1$}  & \text{if}~  $0.4 \leq |d| \leq 2$\\
     \vspace{0.2cm}
     & (Region-3)\\
     \vspace{0cm}
     $  -(T_{\max} - T_{\text{inside}})$  & \text{if}~   $|d| > 2 $\\
\end{tabular}
\right.
\end{align*}

The reward function is divided into four regions based on the value of $d$, which denotes the deviation along the x-axis for the pitch control case and the deviation along the y-axis for the roll control case. The reward function is normalized in the range $[0, -1]$. {Region-1} is where the UAV is within 0.1 m of the target location. Since this is the preferred region for hovering, we impose only a action penalty. Here,  $a_{\text{diff}}$ is the difference between current action value and average of past five action values and $a$ is the current action value. The penalization based on $a_{\text{diff}}$ is to ensure that the control action trajectory is smooth while the penalization on $a$ forces the controller to not select high control values when it is close to the target. In {Region-2}, which is between 0.1 m and 0.4 m from the target, we use a reward function based both on the distance from the target position and the control action.  When $d$ is between 0.4 m and 2 m, which is {Region-3}, we impose the maximum penalization irrespective of the control action and the deviation. The objective is that the UAV needs to get within the 0.4 m mark which is considered as safe zone for landing as quick as possible. We also want the algorithm to accumulate higher rewards (less negative) if it reaches the target point as early as possible and if it stays  in the vicinity of the target. To ensure this, the final part of the reward is designed, where $T_\text{max}$ is the maximum episode time during training and $T_{\text{inside}}$ is the time that the UAV stays within 2 meters from the target. Our training episode ends when $|d|>2$ and this terms kicks in only when a control action pushes the UAV outside of 2 meters.

\begin{figure}[h]
     \centering
     \begin{minipage}[b]{0.23\textwidth}
         \centering
         \includegraphics[width=1.1\textwidth]{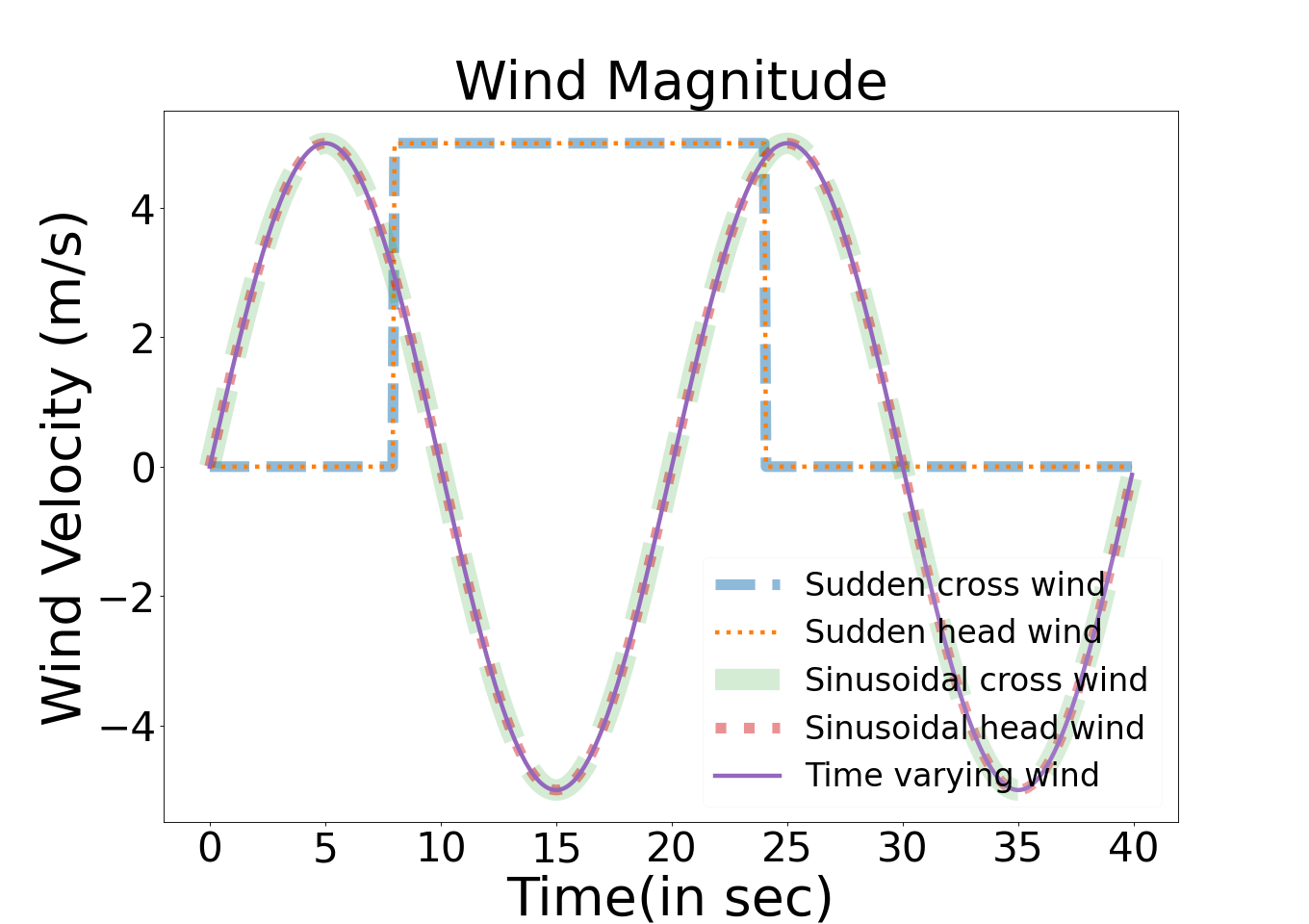}
     \end{minipage}
     \hfill
     \begin{minipage}[b]{0.23\textwidth}
         \centering
         \includegraphics[width=1.1\textwidth]{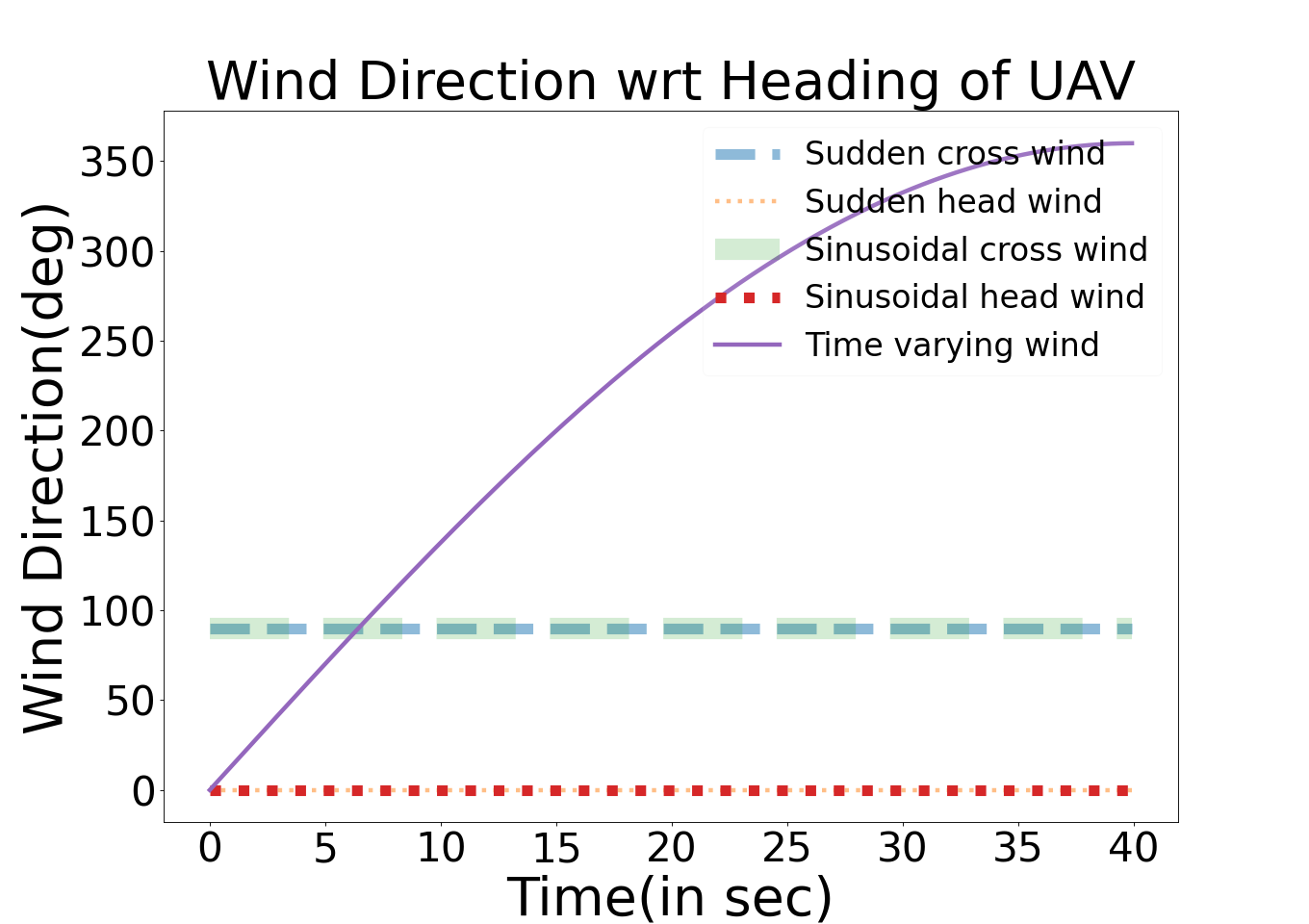}
     \end{minipage}
    \caption{Wind magnitude and direction in different scenarios.}
     \label{fig:wind_scenarios}
\vspace{-0.5cm}
\end{figure}

\begin{figure*}
  \subfloat[Sudden cross wind]{
	\begin{minipage}[b][1\width]{
	   0.23\textwidth}
	   \centering
	   \includegraphics[width=1.1\textwidth]{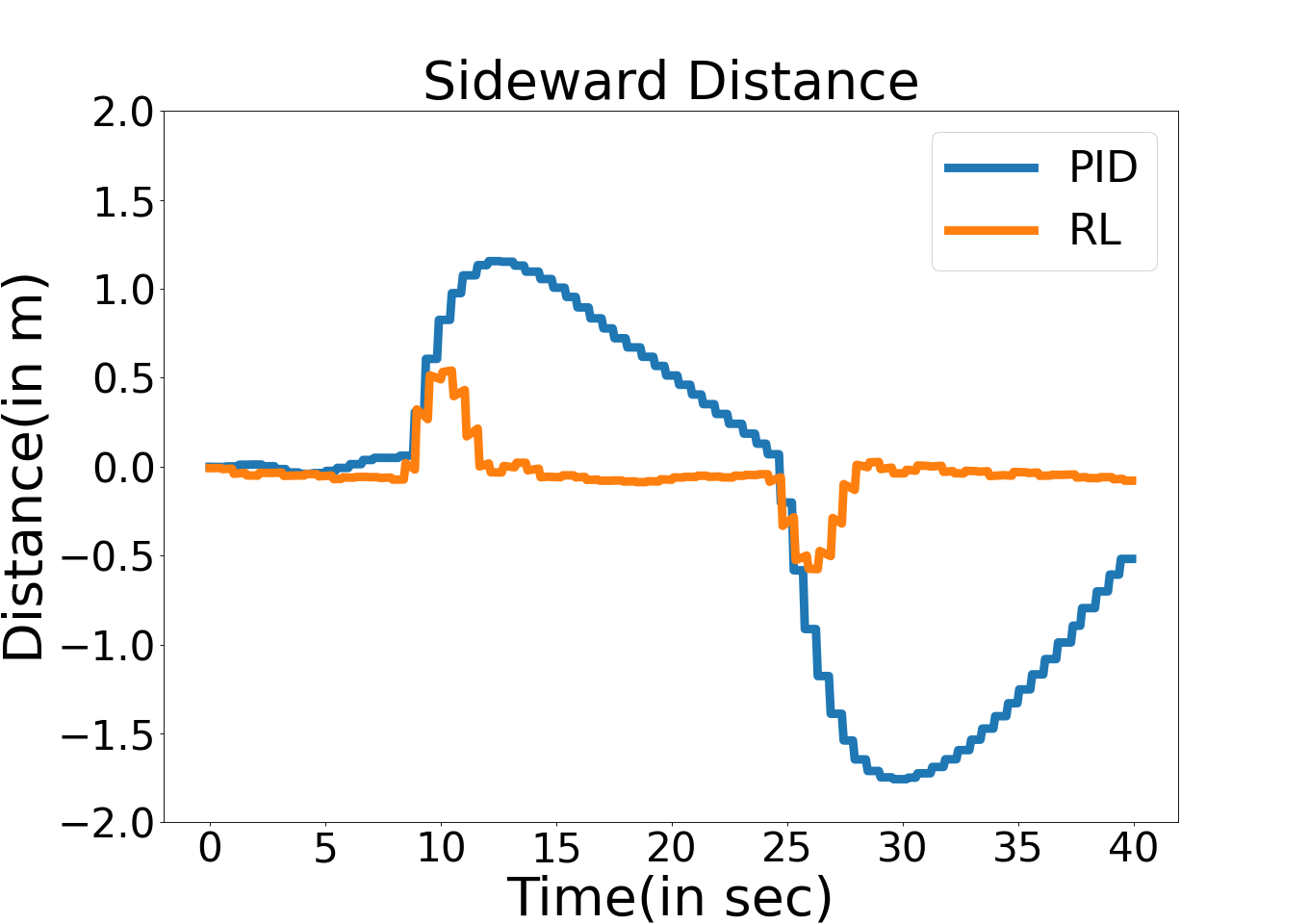}
	   \label{drift_sim_sudcross}
	\end{minipage}}
 \hfill
	\subfloat[Sinusoidal head wind]{
	\begin{minipage}[b][1\width]{
	   0.23\textwidth}
	   \centering
	   \includegraphics[width=1.1\textwidth]{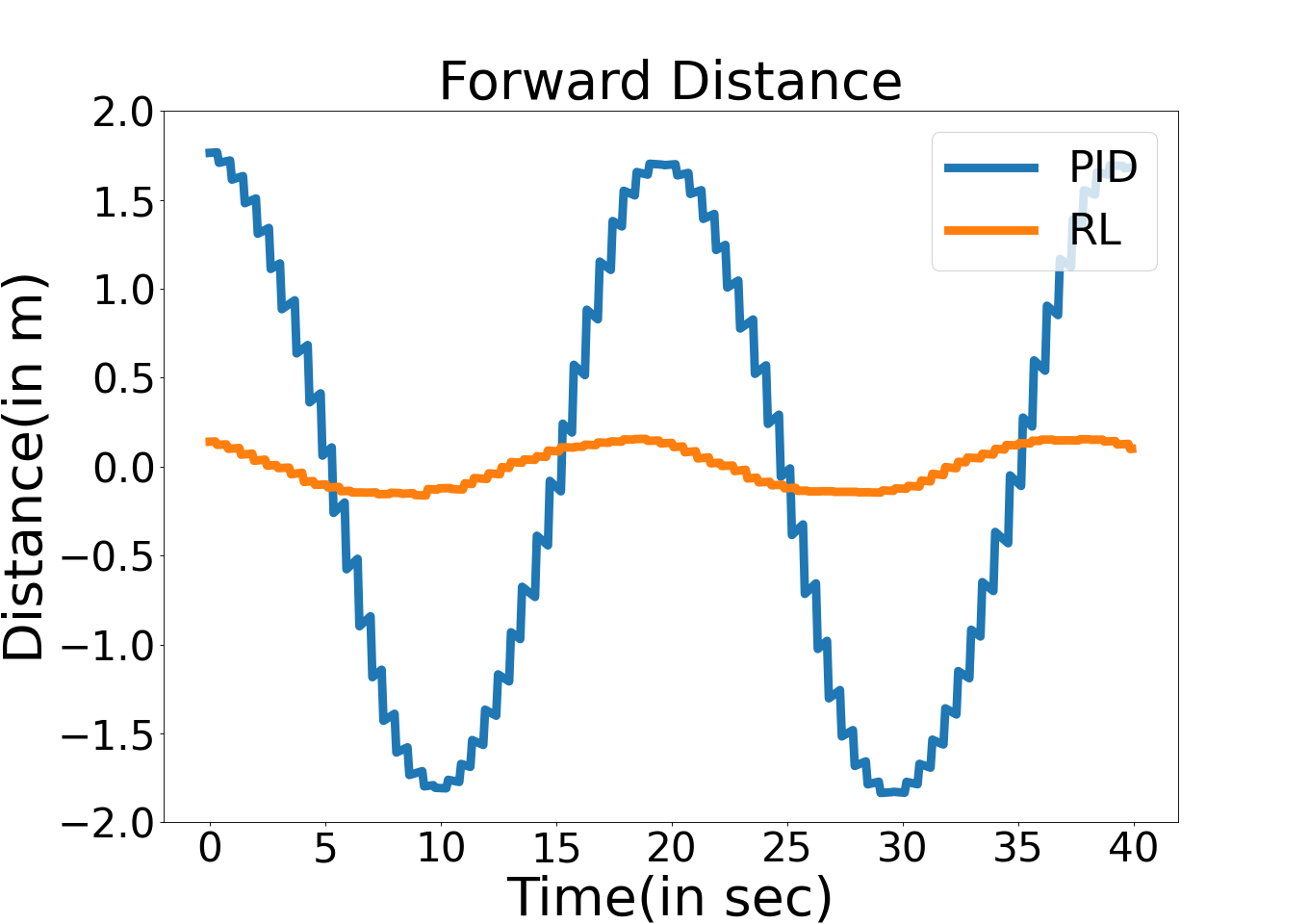}
	   \label{drift_sim_sinhead}
	\end{minipage}}
 \hfill
\subfloat[Time varying wind]{
	\begin{minipage}[b][1\width]{
	   0.23\textwidth}
	   \centering
	   \includegraphics[width=1.1\textwidth]{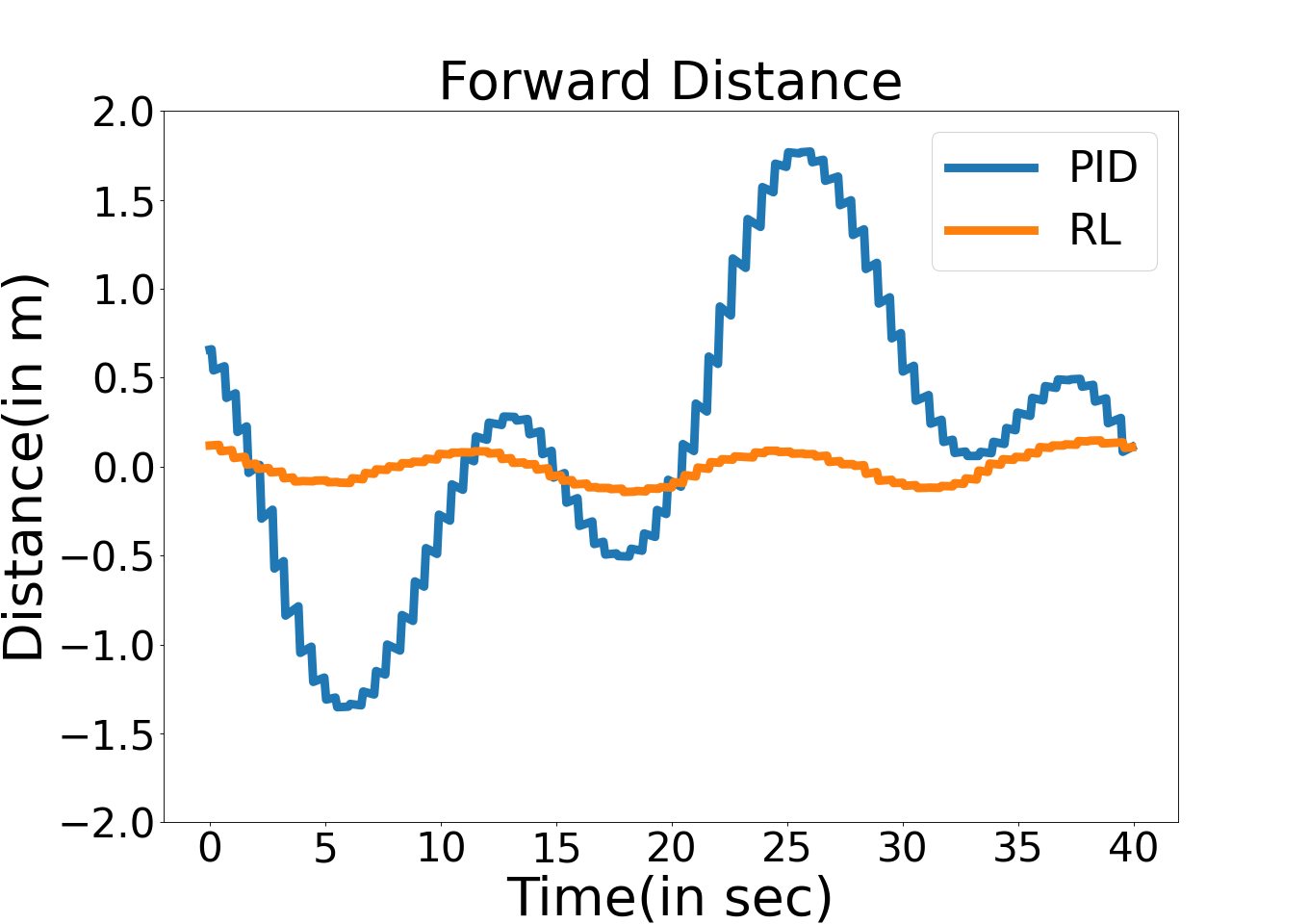}
	   \label{tvw_forward}
	\end{minipage}}
 \hfill
  \subfloat[Time varying wind]{
	\begin{minipage}[b][1\width]{
	   0.23\textwidth}
	   \centering
	   \includegraphics[width=1.1\textwidth]{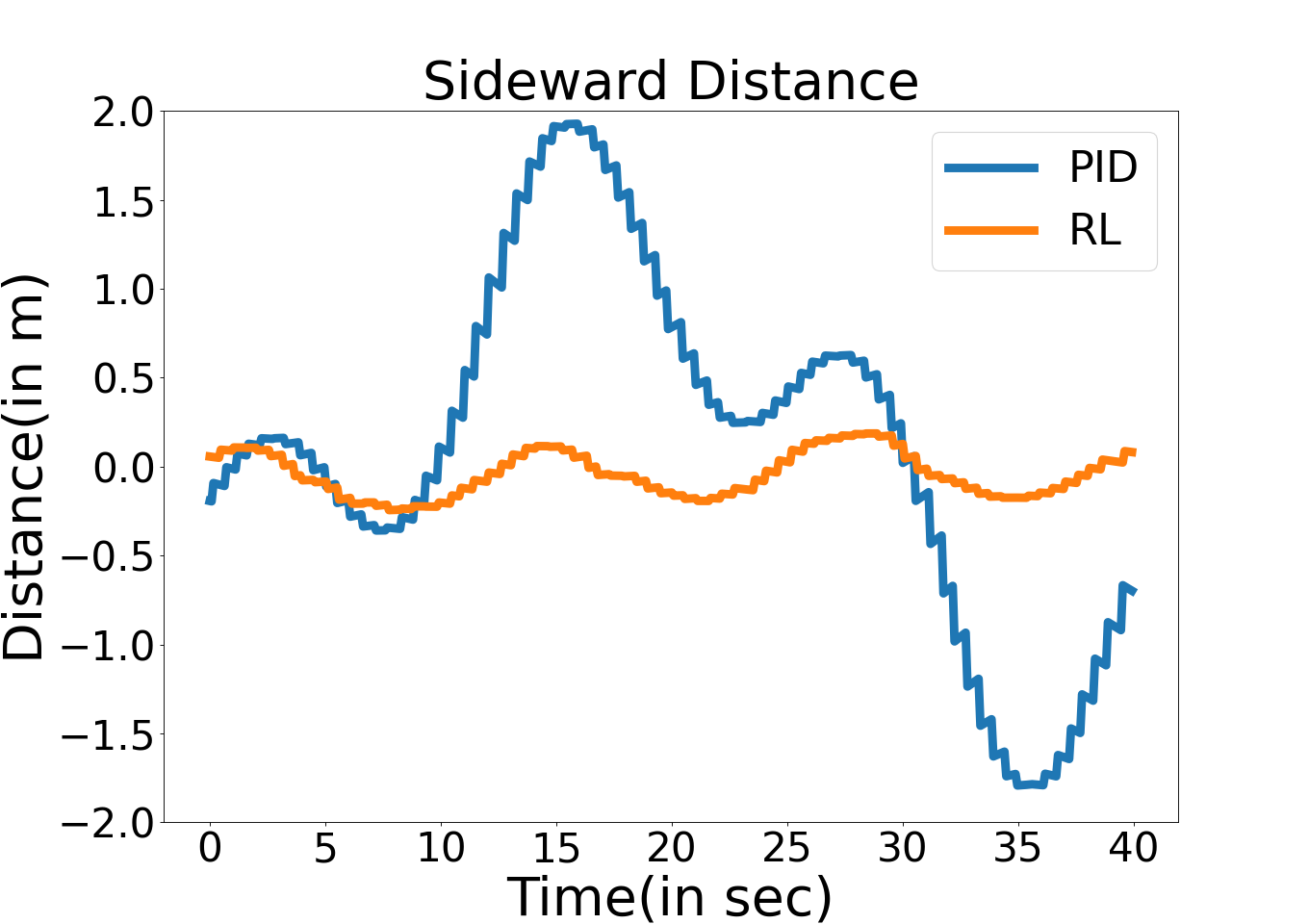}
	   \label{tvw_sideward}
	\end{minipage}}
\caption{Deviation from the desired position in different wind scenarios for the hovering task.}
\vspace{-0.6cm}
\label{drift_sim}
\end{figure*}

\begin{figure*}
\vspace{-0.0cm}
  \subfloat[]{
	\begin{minipage}[b][1\width]{
	   0.23\textwidth}
	   \centering
	   \includegraphics[width=1.1\textwidth]{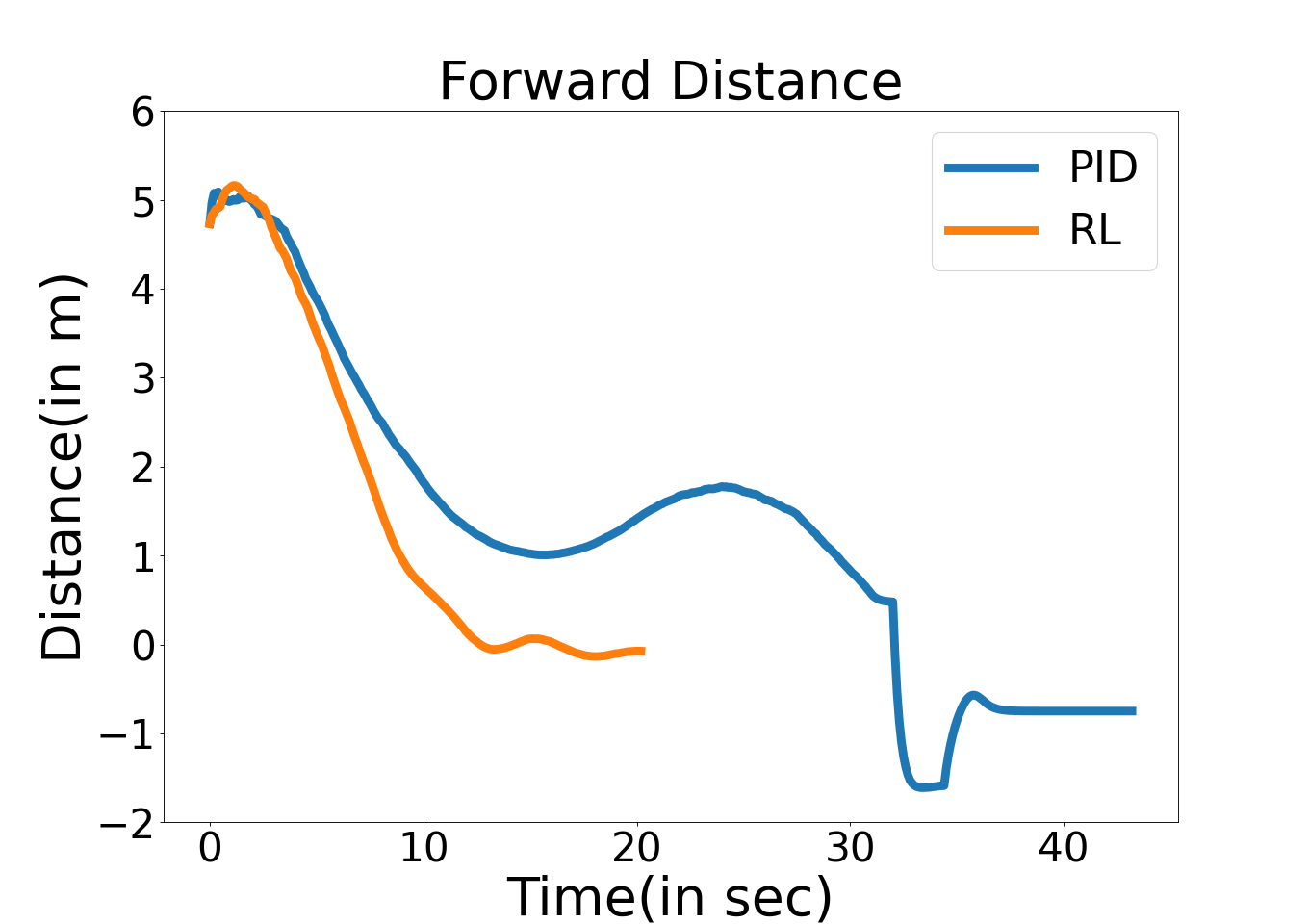}
	   \label{forward_stat_wind}
	\end{minipage}}
 \hfill
  \subfloat[]{
	\begin{minipage}[b][1\width]{
	   0.23\textwidth}
	   \centering
	   \includegraphics[width=1.1\textwidth]{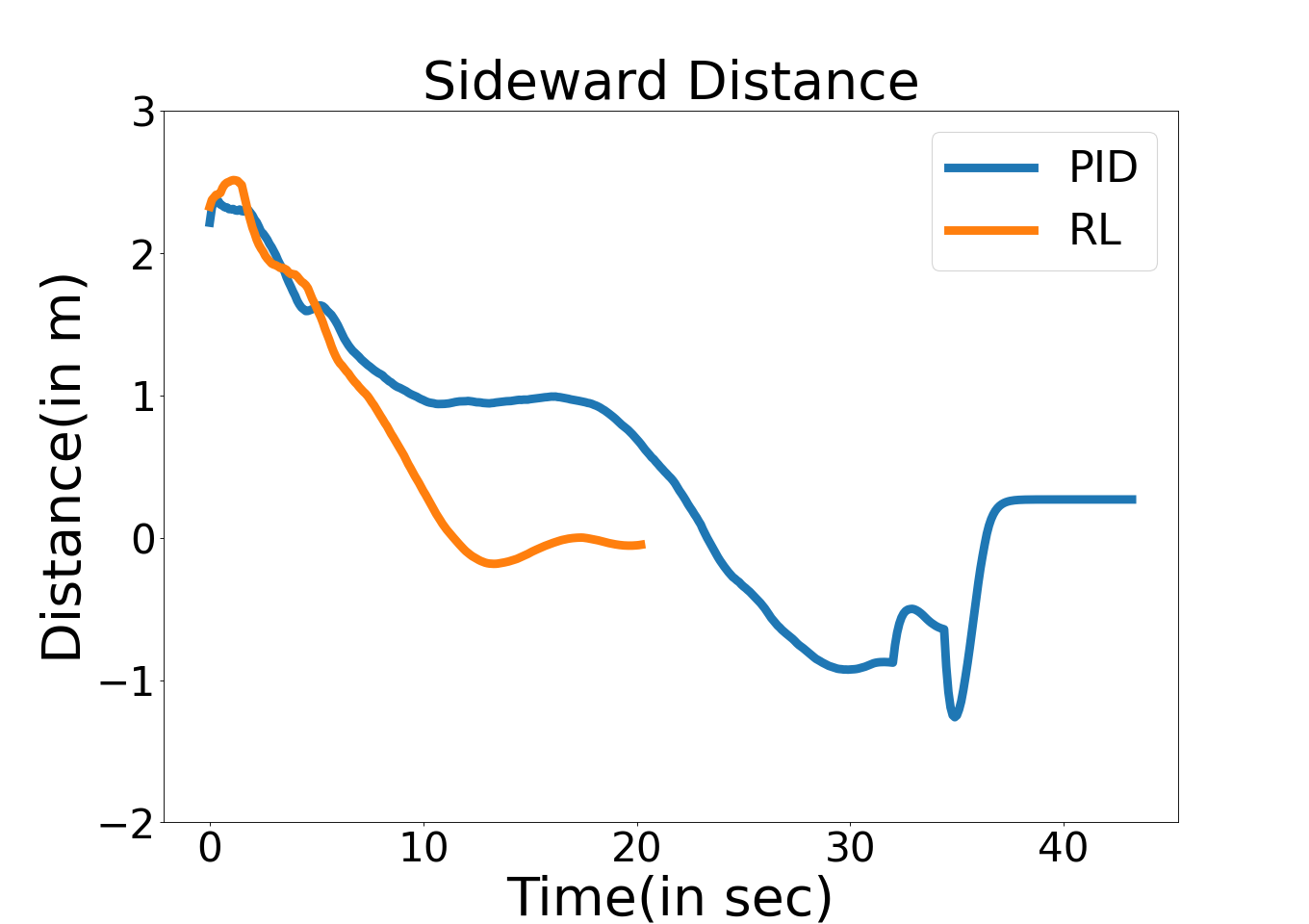}
	   \label{sideward_stat_wind}
	\end{minipage}}
 \hfill
  \subfloat[]{
	\begin{minipage}[b][1\width]{
	   0.23\textwidth}
	   \centering
	   \includegraphics[width=1.1\textwidth]{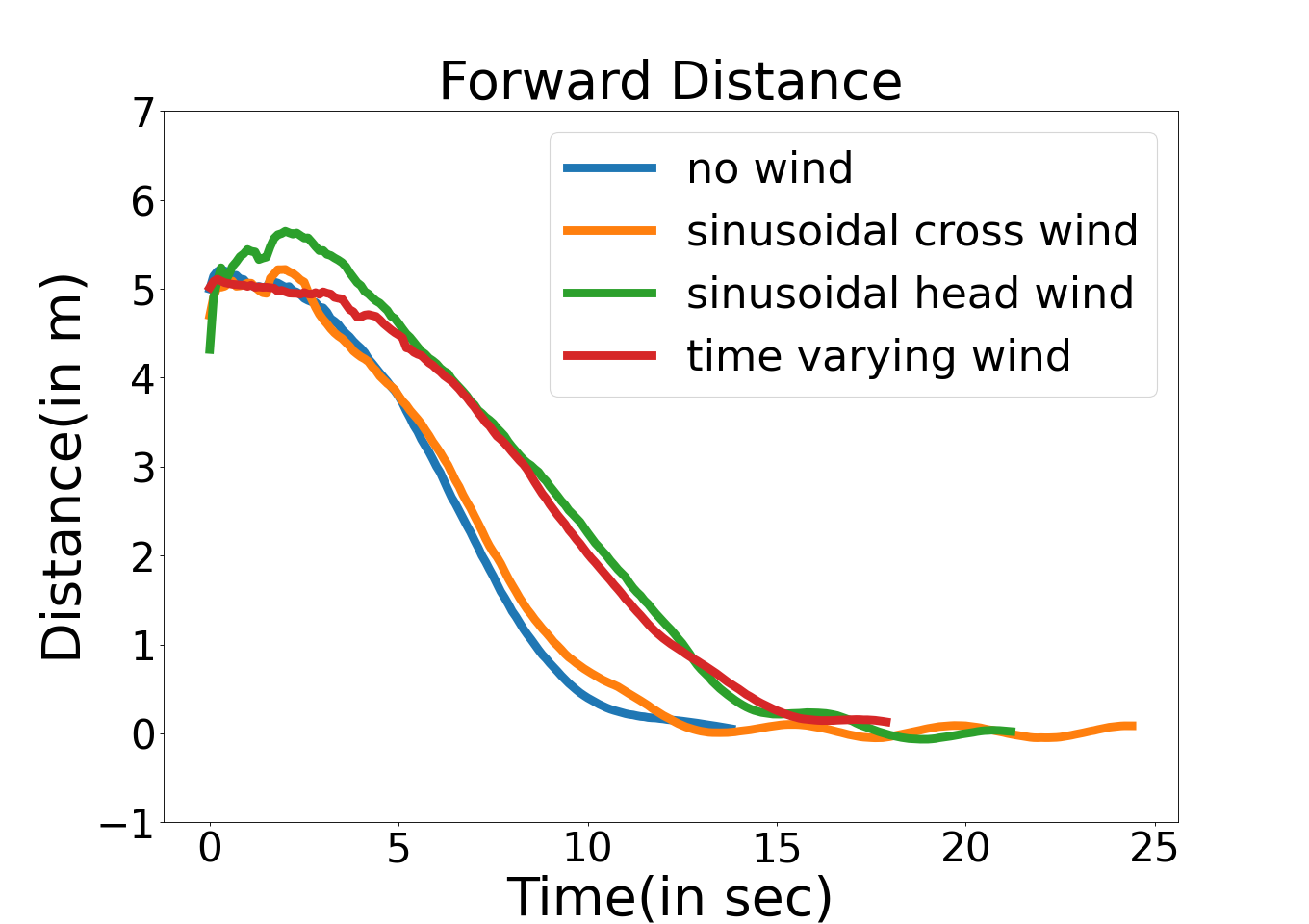}
	   \label{forward_stat_allwind}
	\end{minipage}}
\hfill
	\subfloat[\bluetext{ } ]{
	\begin{minipage}[b][1\width]{
	   0.23\textwidth}
	   \centering
	   \includegraphics[width=1.1\textwidth]{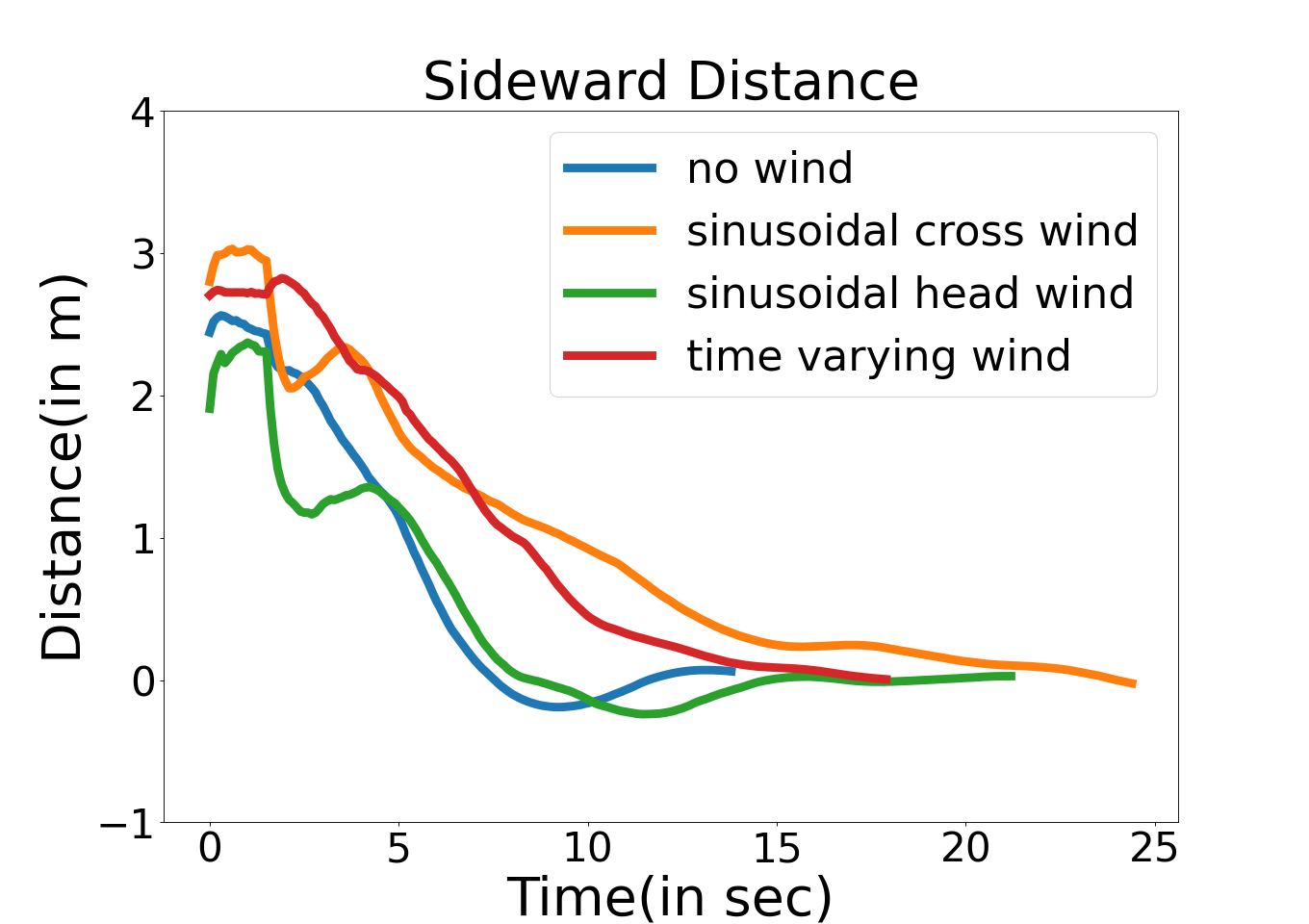}
	   \label{sideward_stat_allwind}
	\end{minipage}}
		
\caption{Distance from the target position in different wind scenarios for the landing task.}
\vspace{-5mm}
\end{figure*}

		

\subsubsection{Robust RL through Domain Randomization} One of the most challenging aspect of RL-based robotic control design is the sim-to-real gap arising from using the control policy trained on a simulator in the real-world. There are inevitable mismatches between the simulator setting and real-world setting in the UAV control problem, not only due to the possible mismatches between the model parameter values of the simulator and real-world UAV, but also due to the presence of adversarial disturbances such as wind gust in the real-world. Domain randomization \cite{sadeghi2017cad, tobin2017domain, peng2018sim} is an approach for overcoming the sim-to-real gap by randomizing the simulator environment during the training of the RL control policy. We adapt the domain randomization approach to develop a UAV controller that is robust against adversarial wind gusts in the real-world environment.

During the RL training, we use the Gazebo simulation engine to generate multiple  wind scenarios: constant magnitude wind (-10 m/s to 10 m/s), sudden wind magnitude change (-5 m/s to 5 m/s magnitude change), and a sinusoidal wind (amplitude of 5 m/s and time periods of 10 secs, 20 secs, 30 secs, 40 secs and 50 secs). We randomize the wind conditions (including its magnitude and directions) over the learning episodes.  It is observed that headwind affects the forward drift and crosswind affects the sideward drift. Hence crosswinds are applied for roll controller training and headwinds are applied for pitch controller training.   

The estimation of relative position and velocity  using the vision system is prone to errors depending on how far the camera is located from the visual cue. So, we  apply the domain randomization to the state estimation system also to be robust against this uncertainty. In addition to this, we also use domain randomization for being robust against issues due to time delays in the vision and control systems.



As mentioned before, we adapt the state-of-the-art TD3 algorithm for training an RL controller in the manner described above. The selected hyper-parameter are: $\gamma = 0.99$, policy delay = 2, learning rate = 1e-4, buffer size = 1e5.  The system used for training is LENOVO Legion Y740-15IRH, which is composed of Intel(R) Core(TM) i7-9750H CPU with 2.60GHz, 2592 Mhz, 6 Cores, and 12 Logical Processors. It features an integrated NVIDIA GeForce GTX 1660 Ti 6GB Graphics and 8GB of LPDDR4 memory with a 128-bit interface. The system has Ubuntu 18.04 OS with Nvidia driver version 440 and CUDA version 10.2. The same system is used for flight simulations and testing.




\section{Simulation Results}

We evaluated the performance of our robust RL controller in the Gazebo simulation environment. We consider the hovering task where the goal is to hover over the target and landing task where the goal is to perform vertical landing on the target, despite the adversarial wind conditions present during these tasks.  The different wind scenarios we considered in our evaluation is shown in Fig.~\ref{fig:wind_scenarios}. As a benchmark, we use a PID control based algorithm developed in our prior work, described in the technical report \cite{lee2022intelligent}. We show that our robust RL controller achieves superior performance compared to this PID controller benchmark.

\subsection{Hovering task in different wind scenarios}

Figure~\ref{drift_sim_sudcross} shows the sideward deviation from the desired position of the UAV in the presence of a sudden cross wind (magnitude changes from 0 to 5 m/s at the 8 second mark). The maximum sideward deviation for our robust RL approach is only one-third of the deviation for benchmark PID controller. Moreover, our  robust RL approach is able to return the UAV to the desired position in less than 2 seconds, while this took at least 15 seconds for the PID controller. Figure~\ref{drift_sim_sinhead} shows the  forward deviations in presence of sinusoidal head wind. The maximum forward deviation for our robust  RL approach is less than 20 cm and it is only one-tenth of the deviation for the benchmark PID control. Figure~\ref{tvw_forward} shows the forward deviation in a more challenging scenario where both the magnitude and direction of the wind change over time. The wind magnitude  is a sinusoidal function with an amplitude of 5 m/s and a time period of 20 seconds. The wind  direction also changes continuously from 0$^{\circ}$ to 360$^{\circ}$ in every 40 seconds. Here also, the maximum deviation for our RL approach is  less than 20 cm and it is only one-tenth of the deviation for the benchmark  PID controller.  Figure~\ref{tvw_sideward} shows the sideward deviation for the same time varying wind scenario.

\subsection{Landing task in  different wind scenarios}


Figures \ref{forward_stat_wind} and \ref{sideward_stat_wind} show the distance of the UAV from the target landing position as a function of time in the presence of a time varying wind scenario described in the above subsection. Note that both the sideward and forward distance converges to zero for our robust RL approach. At the same time, the benchmark PID control approach is not able to make a safe landing in this wind scenario. Figures~\ref{forward_stat_allwind} and ~\ref{sideward_stat_allwind} show that the roboust RL controller comfortably makes landing in a number of wind scenarios. 


\section{Real-World UAV Demonstrations}

\subsection{Experimental Setup}

\begin{figure*}
	\subfloat[Experimental Setup]{
	\begin{minipage}[b][1\width]{
	   0.31\textwidth}
	   \centering
	   \includegraphics[trim=120 -50 0 0,clip,width=\textwidth]{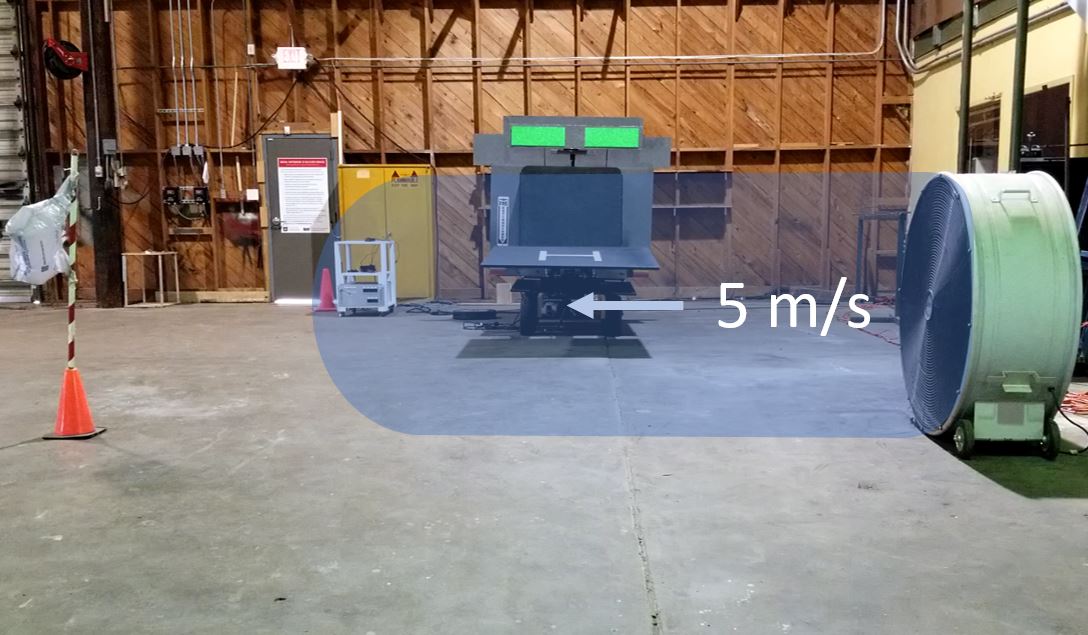}
	   \label{drumfan}
	\end{minipage}}
\hfill
  \subfloat[Forward Distance]{
	\begin{minipage}[b][1\width]{
	   0.31\textwidth}
	   \centering
	   \includegraphics[width=1.1\textwidth]{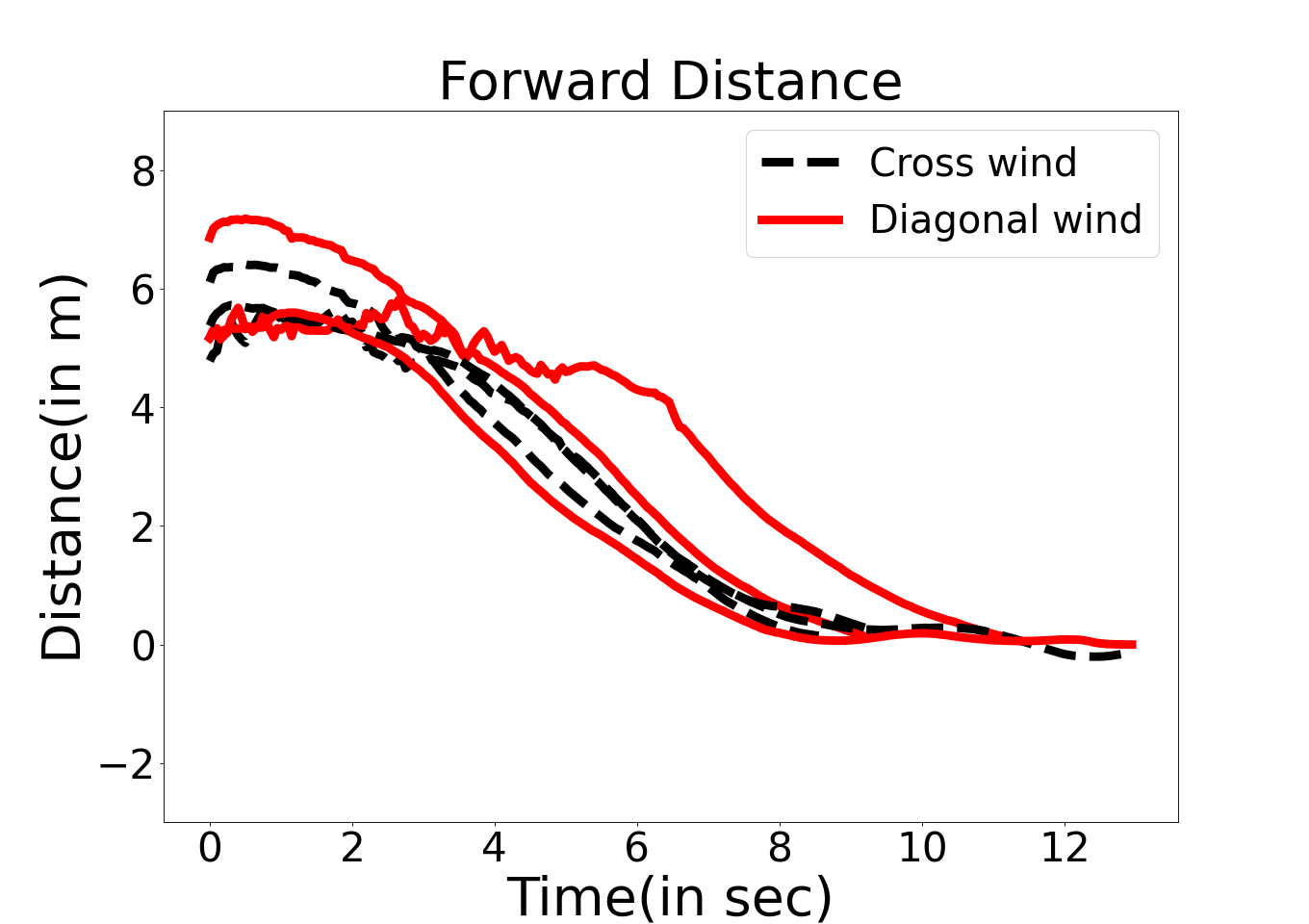}
	   \label{allwind_forward_real}
	\end{minipage}}
 \hfill
  \subfloat[Sideward Distance]{
	\begin{minipage}[b][1\width]{
	   0.31\textwidth}
	   \centering
	   \includegraphics[width=1.1\textwidth]{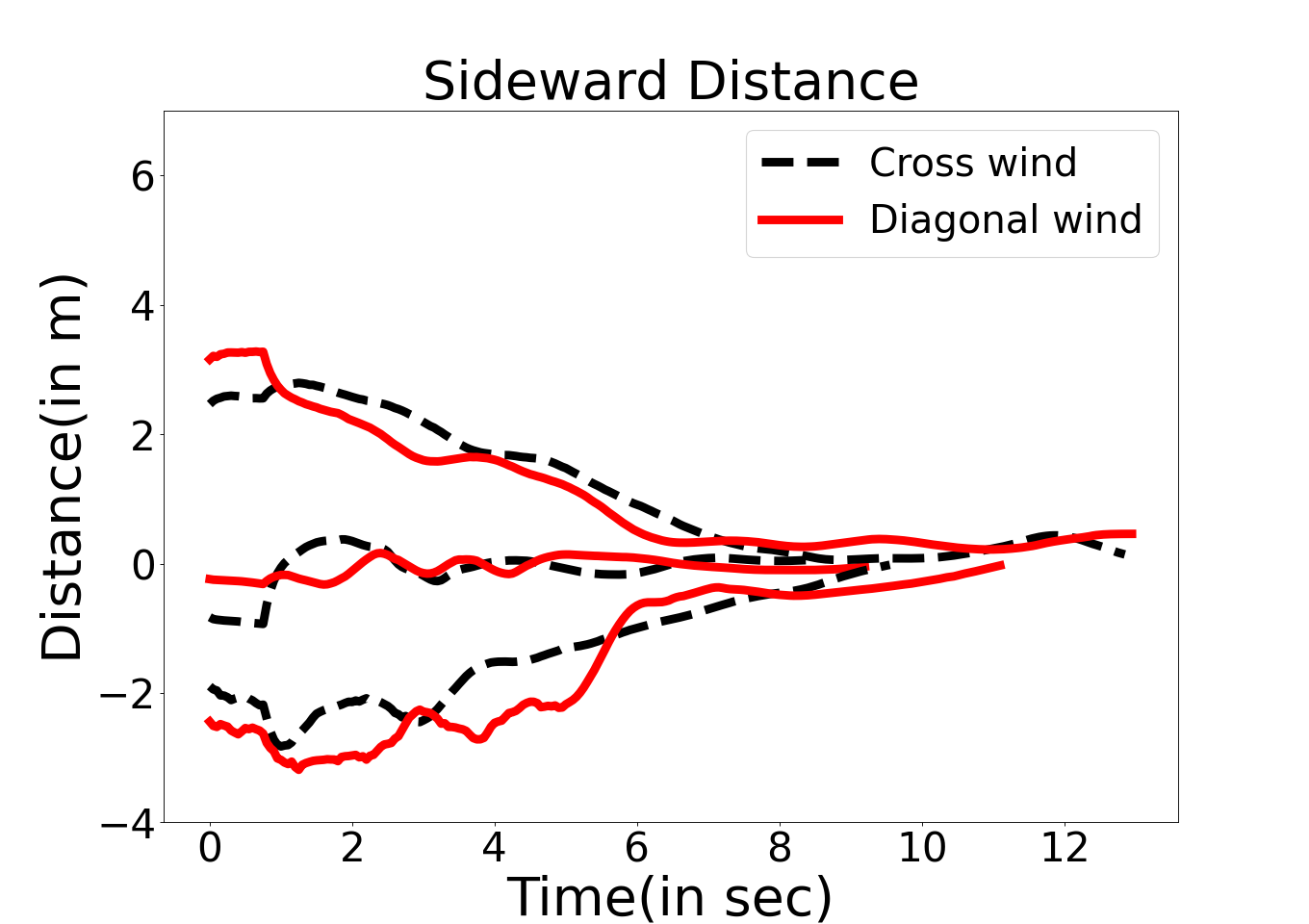}
	   \label{allwind_sideward_real}
	\end{minipage}}
\caption{Landing on stationary platform in different wind scenarios from different initial positions.}
\vspace{-4mm}
\end{figure*}

\begin{figure}[hbt!]
\centering
\includegraphics[width=0.44\textwidth]{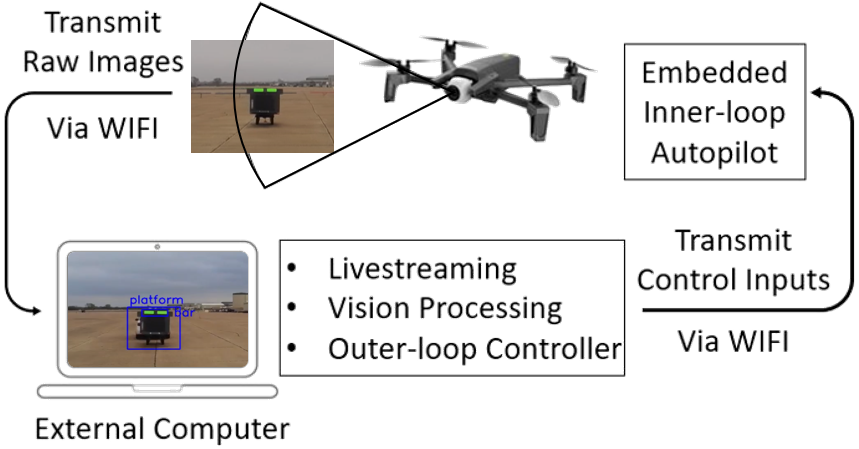}
\caption{Schematic of the UAV control.}
\vspace{-5mm}
\label{fig:process}
\end{figure}

For the real-world demonstration, we use  Parrot Anafi quadrotor UAV \cite{parrot}. The schematic of the control system is shown in Fig.~\ref{fig:process}. The UAV has an embedded (onboard) inner-loop autopilot that controls the rotational speed of each propeller to achieve the commanded inputs generated by the outer-loop RL based control system (offboard).  The UAV is controlled by a Python script that runs on an external computer which communicates with the UAV through the WiFi connection. The UAV transmits raw images captured by the onboard camera to the external computer in real-time, and the computer processes these images to provide estimates of relative position and other state variables. Based on the state estimate, the robust RL controller generates the roll and pitch control actions. The control actions are sent back to the UAV, and  the embedded inner-loop autopilot controls the rotating speed of each propeller based on the received control actions.

We constructed  a sub-scale ship platform model with horizon reference bar and motion deck for our real-world demonstration as shown in Fig.~\ref{fig:demo-model-ship-bar-deck} and Fig.~\ref{drumfan}. The width, height, and length of the ship platform are 5 ft, 5 ft, and 10 ft, respectively. The horizon bar always indicates a perfect horizon, and the motion deck has its own 6 DOF motions in addition to the forward translational motion, which is similar to what would be experienced on a real ship.  We use a drum fan as shown in Fig.~\ref{drumfan} for generating the wind gust. The fan can generate a wind gust with a maximum speed of 3 m/s (in low settings) and 5 m/s (in high settings).


\subsection{Landing task experiment results}

\begin{figure}[hbt!]
\centering
\vspace{-0.0cm}
\includegraphics[width=0.40\textwidth]{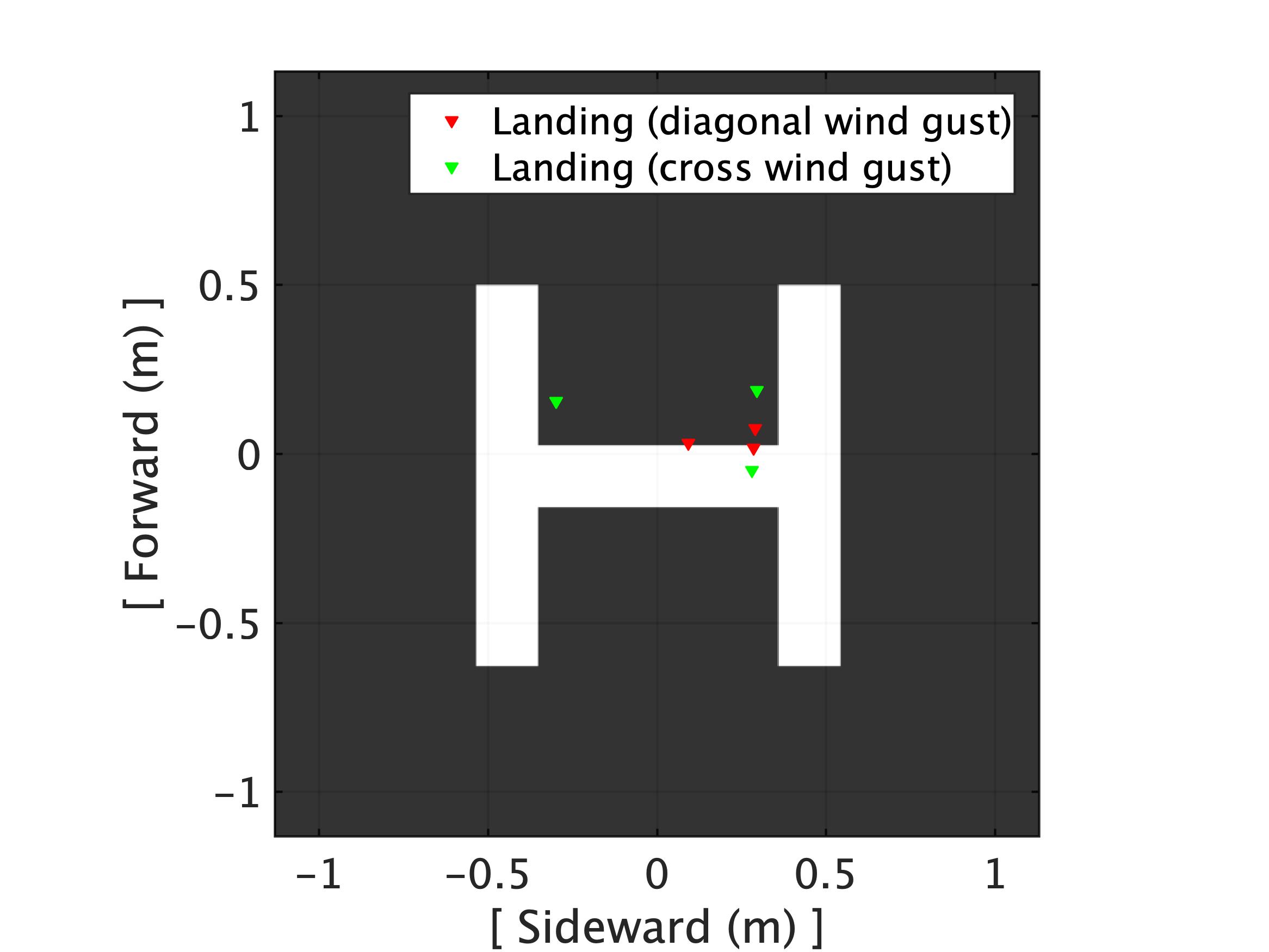}
\caption{Landing points on the platform for the different wind  scenarios.}
\label{landing_rl}
\vspace{-0.5cm}
\end{figure}

In our experiments, the UAV approaches the ship platform from different initial positions. We place the fan at two positions: at 90 degrees with respect to the trajectory of the UAV for generating a cross wind (as shown in Fig.~\ref{drumfan}) and at  45 degrees with respect to the trajectory of the UAV for generating a diagonal wind (figure omitted due to page limit). The forward and sideward distance of the UAV as it approaches the landing platform is shown in Figs. \ref{allwind_forward_real} and \ref{allwind_sideward_real} respectively. Figure~\ref{landing_rl} shows the landing spots on the platform for the same testing scenarios shown in  Figs. \ref{allwind_forward_real} and \ref{allwind_sideward_real}. As clear from these figures, our robust RL approach is able to land the UAV safely and smoothly on the platform in all scenarios.  

\textcolor{blue}{The video of the real-world experiments and demonstrations is available at this} \href{https://www.youtube.com/watch?v=4SiSVvzDrjg}{URL}.


\section{Conclusion}

In this work, we addressed the problem of developing an algorithm for autonomous ship landing for VTOL capable UAVs in the presence of adversarial environmental conditions such as wind gusts. We first developed a computer vision based algorithm which uses the image stream from a single monocular vision camera in the UAV for estimating the relative position of the UAV w.r.t. a horizon reference bar on the landing platform. We then developed a problem specific robust RL algorithm by adapting the domain randomization approach that is capable of controlling the UAV to the landing platform even in adversarial wind conditions.  We demonstrated the superior performance of our approach with respect to a benchmark  PID control based approach by doing evaluations in simulations settings  and in real-world settings.  In our future, we plan to develop a robust meta-learning algorithm that can adapt to more diverse wind scenarios in real-time without compromising on the performance.

\bibliographystyle{IEEEtran}
\bibliography{VTOL-RL-References}

\end{document}